\pdfoutput=1

\documentclass[11pt]{article}

\usepackage[preprint]{acl}
\usepackage{times}
\usepackage{latexsym}
\usepackage{rotating}
\usepackage[T1]{fontenc}
\usepackage[most]{tcolorbox}
 \usepackage{amsmath}
\usepackage[utf8]{inputenc}
\usepackage{booktabs}
\usepackage{multirow}

\usepackage{microtype}
\usepackage{threeparttable}
\usepackage{inconsolata}
\usepackage{hyperref}
\usepackage{graphicx}

\newtcolorbox{storybox}[1][]{
  colback=gray!5!white, 
  colframe=gray!75!black, 
  title=\textbf{#1}, 
  fonttitle=\bfseries,
  sharp corners=south, 
  rounded corners=north,
  boxrule=0.5pt, 
}

\usepackage{todonotes}

\title{Children’s English Reading Story Generation via Supervised Fine-Tuning of Compact LLMs with Controllable Difficulty and Safety}

\author{Qian Shen \\ 
        University of Florida \\
        \texttt{qian.shen1@ufl.edu} \And
        Fanghua Cao\textsuperscript{*} \\
        University of Florida \\
        \texttt{cao.fanghua@ufl.edu} \And
        Min Yao\textsuperscript{*} \\
        University of Florida \\
        \texttt{min.yao@ufl.edu} \AND
        Shlok Gilda \\
        University of Florida \\
        \texttt{shlokgilda@ufl.edu} \And
        Bonnie J. Dorr \\
        University of Florida \\
        \texttt{bonniejdorr@ufl.edu} \And
        Walter L. Leite \\
        University of Florida \\
        \texttt{leitewl@ufl.edu} \\
        \textsuperscript{*}Equal contribution}

\begin{document}
\maketitle

\begin{abstract}
Large Language Models (LLMs) are widely applied in educational practices, such as for generating children's stories. However, the generated stories are often too difficult for children to read, and the operational cost of LLMs hinders their widespread adoption in educational settings. We used an existing expert-designed children's reading curriculum and its corresponding generated stories from GPT-4o and Llama 3.3 70B to design different experiments for fine-tuning three 8B-parameter LLMs, which then generated new English reading stories that were subjected to quantitative and qualitative evaluation. Our method prioritizes controllability over scale, enabling educators to target reading levels and error patterns with a compact, affordable model. Our evaluation results show that with appropriate fine-tuning designs, children's English reading stories generated by 8B LLMs perform better on difficulty-related metrics than those from zero-shot GPT-4o and Llama 3.3 70B, with almost no discernible safety issues. Such fine-tuned LLMs could be more broadly used by teachers, parents, and children in classrooms and at home to generate engaging English reading stories with children's interests, controllable difficulty and safety.
\end{abstract}

\section{Introduction}
Children’s reading fluency is a key marker of reading development and profoundly influences language comprehension, expressive ability, and academic achievement \citep{spear2006children, veenendaal2015oral}. For young learners, varied forms of repeated reading have been shown to effectively boost oral reading fluency \citep{rasinski2016alternative}. Consequently, empirical work indicates that providing diverse, engaging materials elevates children’s interest and motivates more practice \citep{merisuo2014interesting}. Yet it is clearly difficult for teachers and researchers to manually assemble large volumes of content tailored to each child’s preferences. Generative AI offers a promising way to meet this need.

Today, generative AI can be trained to craft engaging, age-appropriate narratives that reflect diverse traditions and to independently produce imaginative text suited for advanced image-generation systems, enriching the storytelling experience \citep{lesner2024ai}. Such stories can resonate with children’s imaginative worlds while offering educational and entertaining narratives aligned with the moral values appropriate to different age groups \citep{leite2025storiza}. Creating stories tailored to children’s interests and needs may enhance motivation, increase reading engagement, and help achieve the goal of improving reading skills \citep{al2024using}.

Prior work \cite{leite2025storiza} used zero-shot prompting based on expert-authored curricula to generate children's stories using OpenAI's GPT-4o and Meta's Llama-3.3 70B \citep{leite2025storiza}. Evaluated against pedagogical theories and NLP metrics, the results revealed a critical limitation: even state-of-the-art closed-source models like GPT-4o struggle with strict pedagogical constraint satisfaction. Specifically, they frequently fail to reliably restrict lexical and syntactic difficulty to target lower grade levels (e.g., K-2 readability scores) without degrading narrative coherence. Furthermore, the recurring API costs of closed models and the massive hardware requirements to deploy 70B+ models locally present prohibitive barriers to widespread educational adoption \citep{irugalbandara2024scaling}.

To democratize access to personalized educational content, deploying compact, local LLMs (under 10B parameters) is an intuitive solution. However, adapting sub-10B models for this task is not merely a standard Supervised Fine-Tuning (SFT) problem; it presents a unique scientific challenge. Compact models typically suffer from a severe trade-off between narrative creativity and strict rule-following. When conditioned on multi-dimensional pedagogical constraints (e.g., strict vocabulary boundaries, syntax simplification, and safety guardrails), small models are prone to mode collapse or logical fragmentation. Standard SFT alone is insufficient to bridge this "controllability gap."

To address this gap, we present a systematic empirical comparison of various fine-tuning and inference strategies for phoneme-controlled story generation. By utilizing high-quality stories, we evaluate how effectively sub-10B models can internalize strict pedagogical alignment through techniques such as standard SFT, reward-weighted SFT, and augmented input. Our primary objective is to provide empirical evidence that, with appropriate training pipelines, compact models can achieve controllable difficulty and safety that match or even surpass those of massive zero-shot 70B models. This empirical foundation demonstrates the viability of broader, equitable adoption of AI-driven reading interventions irrespective of economic or hardware constraints.

Accordingly, our research is guided by two questions: (RQ1) Which SFT strategies most effectively enable sub-10B LLMs to generate children's stories with controllable reading difficulty that meets K–2 pedagogical standards? (RQ2) Can these compact fine-tuned models achieve levels of content safety comparable to those of state-of-the-art zero-shot 70B models?

\section{Related Work}
We surveyed the current research works on the applications of generative AI, controllable story generation, and supervised fine-tuning in educational practice to identify available methods and pinpoint the research gap.

\subsection{AI-Generated Stories for Children}
As generative AI continues to be applied in various fields, researchers have 
explored using multimodal generative AI, spanning text, image, and voice 
generation, to construct children's story generation systems \citep{chowdhury2025large, liu2025cogent}. For example, 
\citet{han2023design} propose AIStory, a visual narrative prototype for 
children's creative expression and literacy, while \citet{fan2025words} develop 
StoryPrompt to enable child–AI co-creation of stories and comics.

Despite these advances, studies reveal persistent limitations. 
\citet{sun2024exploring} find that AI-based storytelling still falls short of 
parents' expectations due to interaction and algorithmic challenges. 
\citet{jin2025they} highlight that consistency between stories and illustrations, 
accuracy relative to the real world, and clear emotional expression are crucial 
for educational use, while content safety remains a fundamental requirement. 
\citet{liao2025ai} further points out that AI-generated stories often contain 
language too complex for young children. In summary, as LLMs are increasingly 
deployed for children's story generation, age-appropriate readability,  and content safety have become critical concerns that the field has yet to 
sufficiently address.

\subsection{Automatic Story Generation with Controllable Difficulty}
Recent advances in LLMs and SFT technologies have enabled clearer control over text readability \citep{glandorf2024towards,kim2025readctrl}, yet a critical challenge remains in educational contexts: the tension between strict early-elementary readability constraints and narrative diversity. Rigid constraints such as those required for phonics curricula often force language models into repetitive, formulaic patterns 
\citep{holtzman2019curious}, a problem particularly pronounced in compact local 
models with limited capacity to balance linguistic creativity with vocabulary 
restrictions \citep{farajidizaji2024possible}. Addressing this trade-off requires 
embedding pedagogical constraints directly into model parameters, motivating 
the use of SFT in our work.

\subsection{Application of Supervised Fine-Tuning in Educational Practice}
SFT has been widely adopted to adapt LLMs to concrete educational tasks. At 
the assessment level, \citet{duongtrung2024bloomllm} propose BloomLLM, which 
fine-tunes an LLM on question–answer pairs labeled with Bloom's taxonomy, 
enabling controllable generation across cognitive levels. In student support, 
\citet{assayed2024transformer} fine-tune a BERT-based model on domain-specific 
advising data, improving contextual accuracy over untuned baselines. More 
broadly, \citet{chen2023finetuning} report that fine-tuned LLMs outperform 
prompt-only models on tasks such as explanation generation and content 
recommendation. SFT has also been integrated into interactive learning systems: 
\citet{gao2025finetuned} fine-tune an LLM within Tailor-Mind to support 
self-regulated learning, with user studies showing improved goal setting and 
monitoring. Building on the general paradigm of aligning pre-trained models to 
human intent through SFT \citep{ouyang2022training}, these works collectively 
suggest that SFT offers a practical route to embed pedagogical structure, 
domain specificity, and safety constraints directly into model parameters.

\section{Methodology}
\begin{figure*}[t]
    \centering
    \includegraphics[width=0.85\linewidth]{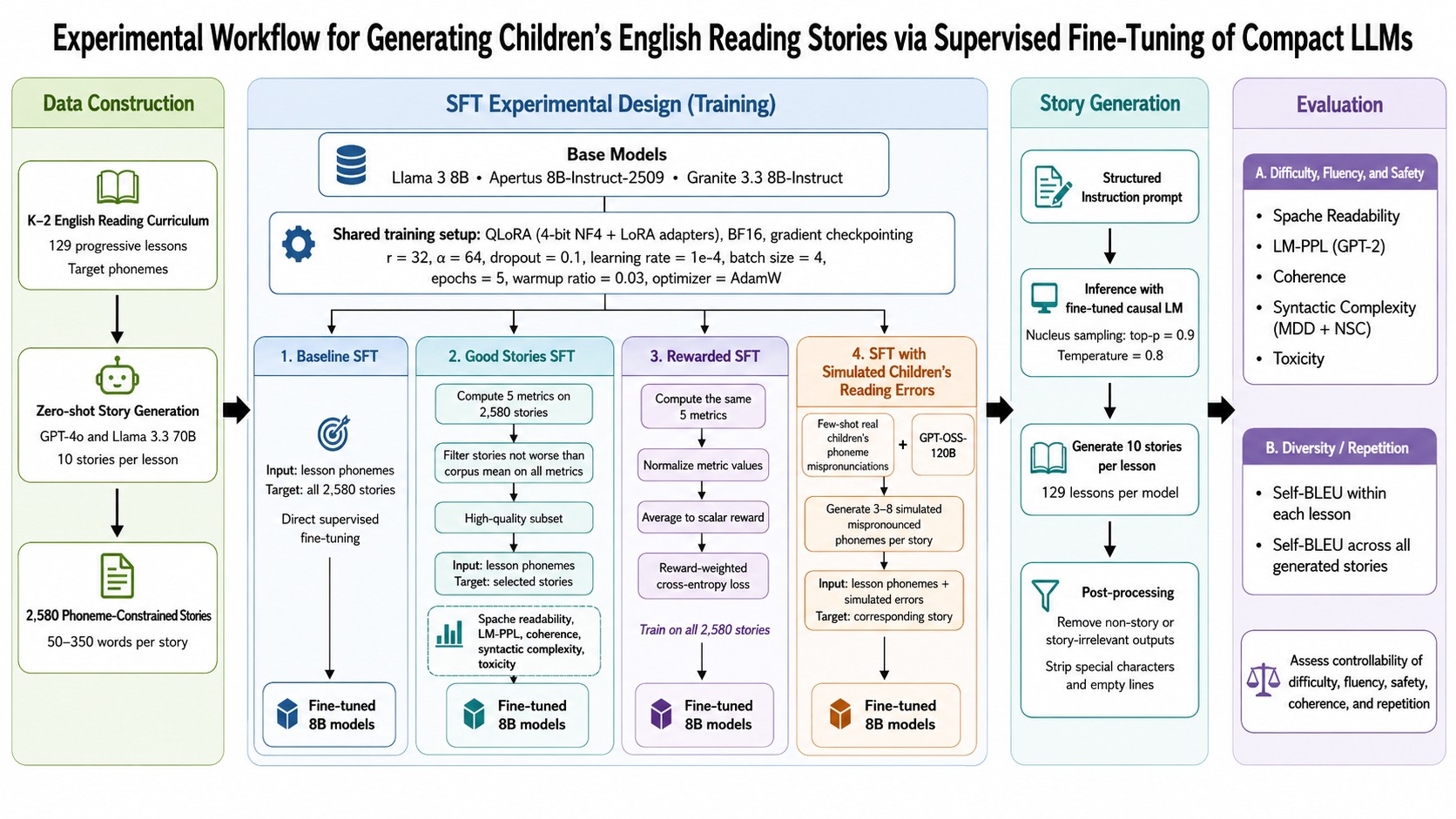}
    \caption{System architecture and experimental workflow for generating children's English reading stories via supervised fine-tuning of compact LLMs.}
    \label{fig1}
\end{figure*}

Through various data preprocessing methods, we constructed multiple SFT experiments on three 8B LLMs using 2,580 stories generated from GPT-4o and Llama-3.3-70B, and generated new English reading stories for children with fine-tuned LLMs. Then, we selected several evaluation metrics to analyze the controllability of story difficulty and safety, and other metrics such as the similarity between stories generated in each experiment. Figure \ref{fig1} and the following sections detail our experiments, reasoning, and evaluations.

\subsection{Data}
Our data derive from a K–2 English reading curriculum from the University of Florida Literacy Institute (UFLI) \cite{lane2022ufli}: 129 progressively difficult lessons, each specifying target phonemes. Figure \ref{fig:curriculum_example} in Appendix B shows an example of this curriculum. With zero-shot prompting, GPT-4o and Llama-3.3 70B were used to generate 10 stories per lesson constrained to the designated phonemes, yielding 2,580 stories \cite{leite2025storiza}. Although these models struggle with strict pedagogical constraint satisfaction in zero-shot settings, their outputs still provide sufficient linguistic quality and narrative coherence to serve as training targets; the pedagogical alignment is then refined through the SFT process. Each story contains 50–350 words, with word counts increasing with lesson difficulty.

\subsection{Supervised Fine-Tuning Experiments}
We established a baseline method using direct SFT on LLMs with our data, and designed three SFT experiments based on different data preprocessing methods.
\subsubsection{Model and Parameter Setting}
In our SFT experiments, we fine-tuned three LLMs: Llama 3 8B, Apertus 8B-instruct-2509, and Granite 3.3 8b-instruct. To adapt the base LLM efficiently, we used  QLoRA \citep{dettmers2023qlora} (4-bit NF4 quantization + LoRA adapters) for parameter-efficient fine-tuning. The LLMs were quantized into 4-bit NormalFloat (NF4) format with double quantization and trained using LoRA adapters (rank = 32, $\alpha$ = 64, dropout = 0.1, learning rate = 1e-4, batch size = 4, epochs = 5, warmup ratio = 0.03, and the optimizer is AdamW) attached to attention and MLP projection layers. Training is performed in BF16 precision with gradient checkpointing to reduce the memory footprint.
\subsubsection{SFT Experimental Design}

\begin{itemize}\setlength{\itemsep}{0pt}
    \item Baseline: We adopt the simple SFT experiment as our baseline. Specifically, we construct input–target pairs by using the phonemes contained in the corresponding curriculum lessons as inputs and the 2,580 previously generated stories as targets, and fine-tune three selected LLMs separately with QLoRA. We then use the fine-tuned LLMs to generate 10 stories for each lesson.
    \item "Good Stories": Some research works on fine-tuning LLMs believe that using higher-quality data can outperform using larger amounts of lower-quality data \citep{chen2023alpagasus,zhou2023lima}. Accordingly, we first computed five metrics for the 2,580 stories previously generated by GPT-4o and Llama 3.3 70B (we detail their computation later): Spache readability, language model perplexity (LM-PPL), coherence, syntactic complexity, and toxicity.To avoid an overly small training set, we selected stories that were not worse than the corpus means on all metrics, yielding 996 stories in total, and used them to fine-tune the three LLMs, constructing input–target pairs from the corresponding phonemes and stories for SFT.
    \item Rewarded SFT: We considered Reinforcement Learning from Human Feedback (RLHF) \citep{bai2022training} as a fine-tuning approach, but our sample size was insufficient to train an effective reinforcement learning (RL) model. Thus, we follow LLM alignment approaches that reinterpret RL as supervised learning with reward-dependent weights \citep{peters2006reinforcement,mukherjee2025offline}. Concretely, each training example is assigned a scalar reward based on the five metrics we mentioned, and the standard cross-entropy loss in the SFT Trainer would be reweighted accordingly. It is closely related to reward-weighted regression schemes explored for offline LLM alignment and self-improvement \citep{qu2024recursive}, and fits within the broader “learning from rewards” framework for LLMs \citep{zhang2024reward}. As in our baseline, we construct input–target pairs by using the phonemes and all the 2,580 stories. During training, we compute a scalar reward $r_i$ by combining five automatic evaluation metrics. Since these metrics are defined on different scales and directions, each metric is first normalized into $[0,1]$. For metrics where lower values indicate better quality (e.g., readability difficulty or perplexity), we apply inverted min–max normalization:
\[
\tilde{m}_i = \max\!\left(0, \min\!\left(1, \frac{b_m - m_i}{b_m}\right)\right),
\]
where $m_i$ is the raw metric value and $b_m$ is a predefined upper bound. For metrics where higher values indicate better quality, we apply standard min–max normalization. The final scalar reward is computed as the unweighted average of the normalized metrics:
\[
r_i = \frac{1}{5}\sum_{k=1}^{5}\tilde{m}_i^{(k)}.
\]
This formulation ensures comparability across metrics while preserving stability during training.

    \item SFT with simulated children's reading errors: Inspired by the Self-Instruct paradigm \citep{wang2023self}, we leverage advanced LLMs to simulate diverse phonemic errors typical of early readers, addressing the data scarcity of real-world error samples. Following the LaMP framework for personalization \citep{salemi2024lamp}, we constructed a few-shot learning setup using data on real children’s phoneme mispronunciations observed while reading several stories, and used OpenAI’s GPT-OSS-120B \citep{agarwal2025gpt} to generate 3–8 simulated child mispronounced phonemes for each of the 2,580 stories. Then, we used the phonemes from each lesson and simulated child mispronounced phonemes as input, and the 2,580 stories as the target to do SFT. This approach aligns with recent findings from "STaR: Bootstrapping Reasoning with Reasoning" \citep{zelikman2022star}, which suggest that training on high-quality, educationally targeted synthetic data significantly enhances model performance in specialized domains.
\end{itemize} 

\subsection{Evaluation Metrics}
To evaluate whether the difficulty and safety of stories generated by fine-tuned LLMs are controllable enough, we employed five evaluation metrics to measure these stories’ difficulty and safety for children. First, we used the Spache Readability score \citep{spache1953new} to assess text difficulty specifically for our target audience of early elementary readers (up to the 3rd grade), where a score below 3.0 shows text at approximately the second-grade reading level.

Second, to measure the fluency and naturalness of the generated stories, we computed LM-PPL. We employed the GPT-2 model via the Hugging Face Evaluate library to calculate the perplexity. 

Third, we introduced a coherence score to quantify local narrative consistency by averaging the number of shared named entities between consecutive sentences, identified using spaCy. Fourth, we evaluated syntactic complexity to capture the structural difficulty of the text. We designed a composite metric that combines the average maximum dependency distance (MDD) within sentences and the average number of subordinate clauses (NSC), computed using spaCy.

Also, text toxicity was assessed to ensure the safety of content intended for children. We employed Detoxify \cite{Detoxify} to identify potential offensive or harmful language, where a score closer to 0 indicates safer content, aiming to filter out stories that exhibit high probabilities of toxic content.

Because the stories generated by GPT-4o and Llama 3.3 exhibit some content-level repetition, we also measured the degree of inter-story repetition among stories produced by the fine-tuned LLMs. We used Self-BLEU \citep{zhu2018texygen} to measure the degree of repetition in the generated stories. It is able to evaluate each story as a hypothesis while treating the remaining stories as references. Thus, higher scores show stronger similarity across outputs and lower generative diversity. To capture repetition at different levels of granularity, we compute Self-BLEU both within each lesson and across the generated stories of each LLM in each experiment.

For a given story belonging to a lesson, we can compute Self-BLEU by using all other stories in the same lesson as references. It reflects how likely the model is to reuse similar expressions, narrative structures, or lexical patterns when conditioned on similar lesson inputs.

In addition to this local view, we measured the repetition between 1,290 stories of each LLM in each experiment by treating every other story as a reference for each story. This provides a broader perspective on whether the model collapses to a narrow set of global templates across lessons. Formal mathematical definitions of all evaluation metrics are provided in Appendix A.

\section{Results}
In our generation stage, we perform inference-only decoding with a fine-tuned causal language model. We apply the LoRA-adapted model parameters to the three base LLMs and generate text autoregressively, conditioned on a structured instruction prompt. Text generation uses nucleus sampling (top-p=0.9) with temperature (0.8), which produces stories without modifying model parameters. Since some outputs from the fine-tuned LLMs contained content extraneous to the stories, and in some cases, the entire output was not a story, we identified these samples through manual inspection, flagging outputs that lacked a coherent narrative structure, consisted mainly of prompt repetition or metadata, or were otherwise not readable as stories. Additionally, to avoid affecting the computation of our story evaluation metrics, we stripped special characters and meaningless blank lines or whitespace from the LLMs' outputs.
\subsection{Quantitative Evaluation}
Table \ref{tab:combined_evaluation1} shows the evaluation results for stories generated by LLMs fine-tuned under Rewarded SFT and Baseline. Examining these results, we find that, except for the overall repetition rate of stories from the Rewarded SFT experiment and the LM-PPL of stories from the Good Stories experiment and SFT with simulated children's reading errors, the metric scores for stories from these three experiments show improvements relative to the baseline. 

Apart from overall repetition, stories from Rewarded SFT and SFT with simulated children's reading errors often approach or surpass those from GPT-4o and Llama 3.3 70B on other evaluation metrics. Across the three 8B LLMs we examined (except for our baseline experiment), stories generated by Llama 3 always underperform those from the other two LLMs on most evaluation metrics. In contrast, the evaluation results of stories produced by Granite 3.3 and Apertus do not show marked differences between the various experimental designs. 

The Spache readability scores of stories generated by our fine-tuned LLMs are mostly lower than those from GPT-4o and are comparable to or even lower than those from Llama 3.3. However, this average Spache readability score still has significant room for improvement to accommodate the reading abilities of children below the second-grade level. The mean toxicity of all our generated stories is very close to 0, suggesting that only a minimal number of stories might pose a safety risk.

In our experiments, the stories generated by LLMs fine-tuned with Rewarded SFT exhibited high stability, being almost entirely free of hallucinations, garbled text, story-irrelevant outputs, or null values. In the other experiments, while the stories from fine-tuned LLMs did contain a small portion of these undesirable outputs, the proportion was less than 10\% of the 1,290 stories generated per model in each experiment.

To further evaluate whether the evaluation metric values for stories generated by LLMs fine-tuned in our SFT experiments differ significantly, in a statistical sense, from those of stories produced by GPT-4o and Llama 3.3 70B, we conducted significance testing. Because the stories are non-identical and our experiments cannot fully eliminate LLM hallucinations or errant responses, although all the stories generated by all the experiments being compared are usable, we used Welch’s T-test to ensure the robustness of the test results. 

Given that the toxicity of the story is nearly zero across the board and that the two metrics related to repetition are not consistently standardized across datasets, we compared the corresponding metric values of fine-tuned LLM stories that achieve the best mean performance on coherence, syntactic complexity, Spache readability, and LM-PPL against those of GPT-4o and Llama 3.3 70B. In the results of Welch’s T-test, all p-values are less than 0.001, and all Cohen's d are greater than 0.8, which shows that the differences in the mean scores for coherence, syntactic complexity, Spache readability and LM-PPL between the stories of fine-tuned LLMs and those of GPT-4o and Llama 3.3 70B are statistically significant.

In summary, quantitative evaluation shows that our SFT experiments, particularly Rewarded SFT, enable fine-tuned 8B LLMs to generate stories with difficulty levels more suitable for K-2 children than those from GPT-4o and Llama 3.3 70B, while presenting almost no discernible safety issues. Complete evaluation results across all four experiments, along with a visual comparison of metric trends across all models and SFT strategies, are provided in Appendix B (Table \ref{tab:combined_evaluation} and Figure \ref{fig3}).

\begin{table*}
\centering
\small
\setlength{\tabcolsep}{4pt}
\caption{Evaluation results of stories generated by LLMs. Values are reported as Mean (SD). $\uparrow$ indicates higher is better; $\downarrow$ indicates lower is better.}
\label{tab:combined_evaluation1}
\begin{tabular}{l c c c c c c}
\toprule
Metric & \multicolumn{3}{c}{Baseline} & \multicolumn{3}{c}{Rewarded SFT} \\
\cmidrule(lr){2-4} \cmidrule(lr){5-7}
 & Llama 3 & Granite 3.3 & Apertus & Llama 3 & Granite 3.3 & Apertus \\
\midrule
Coherence $\uparrow$ & 0.02 (0.05) & 0.07 (0.05) & 0.09 (0.13) & 0.12 (0.14) & 0.18 (0.18) & 0.13 (0.15) \\
Syntactic Complexity $\downarrow$& 4.63 (2.02) & 3.72 (1.05) & 3.41 (0.90) & 3.38 (1.11) & 3.12 (0.88) & 2.96 (0.98) \\
Spache Readability $\downarrow$   & 4.05 (1.04) & 3.52 (0.93) & 2.83 (0.47) & 2.71 (0.70) & 2.56 (0.60) & 2.34 (0.68) \\
Toxicity $\downarrow$             & 0.01 (0.04) & 0.06 (0.12) & 0.00 (0.01) & 0.01 (0.02) & 0.02 (0.06) & 0.02 (0.07) \\
LM-PPL $\downarrow$                & 23.16 (13.22) & 24.91 (10.21) & 16.49 (3.88) & 16.86 (4.29) & 14.55 (3.56) & 19.73 (6.64) \\
Repetition in lessons $\downarrow$& 0.03 (0.03) & 0.11 (0.09) & 0.12 (0.06) & 0.11 (0.06) & 0.12 (0.07) & 0.09 (0.08) \\
Total repetition $\downarrow$     & 0.21 (0.09) & 0.37 (0.16) & 0.40 (0.08) & 0.37 (0.10) & 0.42 (0.11) & 0.32 (0.13) \\
\bottomrule
\end{tabular}
\end{table*}

\subsection{Qualitative Analysis}
In our evaluation of the newly generated stories, we observed outliers in some evaluation metrics. Primarily, the mean LM-PPL values for stories from SFT with simulated children's reading errors were higher than those of other experiments. Observation of these stories revealed that their higher LM-PPL values were either due to their brevity, which led to increased uncertainty and higher cross-entropy per word, or because the LLMs appended content irrelevant to the narrative after generating a short story. This caused an abrupt genre shift, thereby increasing the text's perplexity.

Simultaneously, we observed that some stories in the newly generated sets exhibited toxicity scores significantly higher than others, and we examined these stories closely. We identified several potential factors contributing to the higher toxicity scores, such as the inclusion of language like "you're too fat," "this smells too stinky," and "the pig can gag," or the appearance of characters portrayed as unfriendly or potentially harmful, such as rats. These elements pose a risk of causing discomfort to children or exposing them to content such as body shaming. Overall, the number of stories containing such content was five or fewer in each dataset of 1,290 stories.

Additionally, the instances of excessively high Spache readability grade scores or low coherence scores in some stories were attributed to the presence of difficult words and numerous line breaks, respectively. The syntactic complexity of some stories was collectively elevated by higher rates of subordination, coordination density, complex nominal phrases, premodification, and parentheticals, as well as unconventional word order and fragmented sentence structures.

Figure \ref{fig2} in Appendix B shows a "bad" story and a "good" story from our generated stories as an example. The comparison between the two stories reveals notable differences in linguistic suitability for early children. The “good” story maintains clear coherence, following a straightforward sequence of actions that aligns well with the narrative-processing abilities of young children. Events progress linearly and predictably, and character references remain stable, enabling developing readers to follow the plot without excessive inferential demands. In contrast, the “bad” story introduces a broad and descriptively dense setup, then shifts abruptly toward unrelated or ill-formed fragments. These discontinuities reduce overall coherence and make the narrative less suitable for young readers who benefit from a tighter event structure.

In terms of syntactic complexity, the “good” story employs short, regular sentence structures characteristic of early-grade leveled texts. The “bad” story, by comparison, alternates between long, multi-clause sentences and incomplete constructions, creating variability that may be less aligned with the decoding patterns expected at this developmental stage. The lexical difficulty further highlights this contrast: the “good” story uses high-frequency, concrete vocabulary appropriate for K–2 learners, whereas the “bad” story incorporates abstract, low-frequency terms (“spectacle,” “insatiable,” “peculiar”) that can make comprehension more challenging for beginning readers. These lexical choices also contribute to higher expected perplexity, indicating less predictable and less contextually grounded language. 

Beyond outliers in individual metrics, we also examined the pattern of inter-story repetition reflected in Self-BLEU scores. As shown in Table \ref{tab:combined_evaluation1}, for most generated story datasets (every 1,290 stories per LLM in each experiment), these scores indicate suboptimal diversity. Upon inspection of the generated stories, we observed the recurrence of certain names, animal names, locations, and activities across multiple narratives. Even combinations of these recurring entities, such as Sam, Pam, mats, pigs, and farm, also reappear. Furthermore, some stories feature the same place name repeatedly, albeit attributed to different states within the respective narratives.

Overall, while both stories are grammatically interpretable, the “good” story is more closely aligned with early-elementary reading expectations, whereas the “bad” story exhibits features across coherence, syntax, and vocabulary that render it comparatively less suitable for this age range.

\section{Discussion}
The repetition patterns observed above are likely attributable to the characteristics of the source training data. We believe that these phenomena are related to the source stories from GPT-4o and Llama 3.3 70B themselves. In the stories from Llama 3.3 70B, we also observed similar patterns, such as the recurrence of content involving Sam, Pam, mats, pigs, and farm. This issue arose because we could only use a small number of keywords for batch-generating stories with Llama 3.3 70B due to the limitations of the insufficient data collected at the time. The stories from GPT-4o, which used varied keywords, exhibited lower content repetition; however, they also featured recurring place names, consistently attributed to different states such as Ohio, Illinois, or Kansas. This indicates that after fine-tuning, the generative preferences of these 8B LLMs are strongly influenced by the stories used for the fine-tuning process. It aligns with the findings of \citet{santurkar2023whose} and \citet{chung2024scaling}, who demonstrate that the specific task patterns and behavioral attributes encoded in fine-tuning data directly dictate the model's downstream generation capabilities.

In our evaluation process, while we have utilized quantitative metrics based on NLP theory to assess the stories, feedback, and opinions from experts, teachers, parents, and children are still lacking. Incorporating expert ratings, focus group interviews, and collecting feedback from users of the story-generating LLMs would enable us to better, from the perspectives of data and prompt engineering, guide the LLMs to learn the characteristics of stories that are more suitable for children's needs and to generate new stories more effectively.

In practical classroom applications, stories generated by the fine-tuned 8B LLMs we used are more suitable in difficulty for children from kindergarten to 2nd grade compared to those from the GPT-4o API and the Llama 3.3 70B model, and they present no apparent safety issues. Compared to APIs that require payment and 70B LLMs that demand substantial hardware resources for deployment, both the fine-tuning and inference costs for 8B LLMs are significantly lower, allowing them to be designed as back-end models for websites and software. Teachers, parents, and children can use these fine-tuned, local models for free in the classroom or at home to generate stories suitable for children's reading.

However, 8B LLMs are still too large for deployment on mobile devices or for users who wish to fine-tune the models with their own data for unique pedagogical tasks. Additionally, the previously identified issues within the generated stories, such as some inappropriate content and complex syntax, need to be resolved. Furthermore, our current experimental designs are mutually independent, and the effects of combining these designs remain an area for further exploration.

To ensure the stories generated by these LLMs better meet the needs of teaching and learning, as we have previously mentioned, we still require more stories generated from keywords by teachers, parents, and children, authentic data on children's reading errors, expert ratings of the stories, and any form of user feedback.
\section{Conclusion}
LLMs have been widely utilized in educational practices such as children's story generation. We have used GPT-4o and Llama 3.3 70B to generate English reading stories for children from kindergarten to second grade, based on an expert-authored curriculum. However, the cost and hardware requirements of LLMs hinder their widespread implementation in every classroom. We designed three SFT experiments to fine-tune three popular 8B LLMs and generated new stories based on the same curriculum. We evaluated these stories from both quantitative and qualitative perspectives, and discussed the existing problems and the challenges of applying these fine-tuned LLMs in educational practice.

Our constructed SFT experiments are more effective at generating children's English reading stories with controlled difficulty and safety than the direct SFT baseline. Specifically, stories generated by the 8B LLMs after being fine-tuned in our experiments surpass those from zero-shot GPT-4o and Llama 3.3 70B on difficulty-related evaluation metrics, and they present almost no apparent safety issues. Our "Rewarded SFT" experiment performed comparatively better, generating highly stable content that was nearly devoid of any story-irrelevant material. Among the three 8B LLMs we selected, Llama 3's performance was relatively weaker than the other two. 

Our fine-tuned, compact LLMs will further empower more classrooms and families to freely and conveniently generate English reading stories for children from kindergarten to second grade, ensuring these stories possess controllable reading difficulty and are appropriate in their safety for this age group.

In future research work, we plan to collect more stories generated by people using these LLMs with keywords of interest and combine this with expert ratings and feedback from users of the LLMs to construct a complete RLHF pipeline. This will enable us to use higher-quality data, more human feedback, and more effective LLMs fine-tuning strategies to better fine-tune LLMs for generating stories more suitable for children. Furthermore, we will also consider achieving good story generation performance with smaller LLMs through methods such as model distillation, enabling our fine-tuned LLMs to be used by more teachers, parents, and children in classrooms, at home, on computers, mobile phones, and tablets, regardless of their economic situation or hardware resources.

\section*{Limitations}
In addition to the points already discussed, our research also has some other limitations. First, due to computational resource constraints, we were limited to three L4 GPUs, which prevented us from employing certain SFT strategies in our experimental design; hyperparameters (e.g., the number of training epochs) were also constrained. Second, because related projects were progressing concurrently, we lacked sufficient sample size prior to initiating this study to apply certain fine-tuning techniques, such as completing a full RLHF pipeline. Third, due to IRB review and constraints on practical deployment, our study was built entirely on batch-generated stories and lacks authentic narratives from children, parents, and teachers. Similarly, the generated stories lack expert human evaluation to corroborate the NLP-based automatic evaluation. Moreover, because we used only a single expert-designed curriculum for children’s English reading education, further work is needed to extend our approach to additional curricula and more languages. Furthermore, automated phoneme coverage evaluation remains an open challenge, as standard grapheme-to-phoneme tools struggle with out-of-vocabulary items and irregular orthographic patterns commonly produced by LLMs. We have attempted g2p-en and NLTK but the non-standard curriculum notation (e.g., "CVC words") and large phoneme sets per lesson produced near-zero hit rates. 

Furthermore, in our Rewarded SFT experiment, the scalar rewards were computed using the same automatic metrics later used for evaluation, which raises the possibility that the model was optimized toward the evaluation proxies rather than toward true educational quality. We consider this a limitation and note that human evaluation would be necessary to rule out such over-optimization. Additionally, the comparisons between fine-tuned 8B LLMs and GPT-4o and Llama 3.3 70B are not fully controlled, as the larger models were evaluated only under zero-shot prompting conditions without prompt optimization or few-shot examples. The observed performance differences should therefore be interpreted in this context rather than as a general claim about model capability. Additionally, the fine-tuned models were evaluated on stories generated from the same 129 lessons used during training, rather than on a held-out set of unseen phoneme constraints. This limits our ability to assess generalization to novel lesson inputs.

\section*{Ethics} This work involves the generation of educational content for children and therefore requires particular care regarding data provenance, child safety, and deployment. Our training data are derived from an expert-designed K–2 reading curriculum and from stories synthetically generated by large language models, rather than from children’s personal data or identifiable records. In addition, our use of a small, authentic sample of children's reading errors has been 
approved by the Institutional Review Board of University of Florida 
(Protocol \# IRB202401327). The simulated reading-error inputs used in one experimental condition are synthetic approximations intended to model common early-reader phonemic mistakes, not diagnostic judgments about any individual child. Nevertheless, synthetic data may encode biases, repetitive patterns, or inappropriate language inherited from source models, and our own analysis found a small number of outputs containing potentially harmful expressions, including insults, disgust-related language, and body-shaming content. For this reason, these models should not be deployed as fully autonomous systems for children. Responsible deployment should include human oversight by teachers, parents, or other adult supervisors; automatic safety filtering for toxicity and related harms; conservative blocking or review policies for outputs that include appearance-based insults, humiliation, or other potentially distressing language; and clear disclosure that generated stories may still contain errors or unsuitable content. These systems are best used as assistive tools for draft generation and adaptation, not as unsupervised replacements for educators or caregivers. Future work should strengthen this pipeline through expert human review, broader safety evaluation beyond general toxicity, and governance practices that ensure children’s well-being remains the primary criterion for deployment. In preparing this manuscript, the authors made use of AI-assisted writing tools for language polishing. All scientific content, analyses, and conclusions are the sole responsibility of the authors.
\bibliography{custom}
\newpage
\appendix

\section*{Appendix A: Mathematical Formulas for Evaluation Metrics}
Spache Readability score:

This metric estimates the grade level of a text by combining the average sentence length and the percentage of unfamiliar words according to the following formula:
\begin{equation}
Spache = 0.141 \times \bar{L}_{sent} + 0.086 \times P_{unfamiliar} + 0.839
\end{equation}

LM-PPL:

For a sequence of tokens $X = (x_1, \dots, x_T)$, this is defined as:
\begin{equation}
PPL(X) = \exp \left( -\frac{1}{T} \sum_{i=1}^{T} \log p_\theta(x_i | x_{<i}) \right)
\end{equation}

Coherence Score:

Let $E(S_i)$ be the set of named entities in the $i$-th sentence; the coherence score is derived as:
\begin{equation}
Coh = \frac{1}{N-1} \sum_{i=1}^{N-1} |E(S_i) \cap E(S_{i+1})|
\end{equation}
where $N$ is the total number of sentences and $|\cdot|$ denotes the cardinality of the set.

Syntactic Complexity:
\begin{equation}
Syn = \frac{1}{N} \sum_{i=1}^{N} \left( \text{MDD}_i + \text{NSC}_i \right)
\end{equation}
where $N$ is the total number of sentences, $\text{MDD}_i$ is the maximum 
dependency distance within the $i$-th sentence (i.e., the largest syntactic 
distance between a word and its head across all dependency arcs in that 
sentence), and $\text{NSC}_i$ is the number of subordinate clauses in the 
$i$-th sentence.

Self-BLEU:
\begin{equation}
\text{Self-BLEU}(s_i) = \text{BLEU}\left(s_i,\ \{ s_j \mid j \neq i,\ s_j \in L \}\right).
\end{equation}
where $s_i$ denotes the $i$-th story, $L$ is the set of all stories belonging 
to the same lesson, and the remaining stories $\{s_j \mid j \neq i, s_j \in L\}$ 
serve as references for computing the BLEU score. 

Repetition between 1,290 stories:
\begin{equation}
\text{GlobalSelfBLEU} = \frac{1}{N} \sum_{i=1}^{N} \text{BLEU}(s_i,\ \{ s_j \mid j \neq i \})
\end{equation}
where $N$ is the total number of stories generated by a given model in a given 
experiment, and each story $s_i$ is evaluated against all other stories 
$\{s_j \mid j \neq i\}$ as references.

\section*{Appendix B: Tables and Figures}
In this appendix, we provide additional details regarding our experimental setups, evaluation results of newly generated stories, examples of the K-2 English reading curriculum.

\begin{sidewaystable*}
\centering
\small
\setlength{\tabcolsep}{4pt}
\caption{Evaluation results of stories generated by LLMs for all the experiments, $\uparrow$ indicates higher is better; $\downarrow$ indicates lower is better.}
\label{tab:combined_evaluation}
\begin{tabular}{l l c c c c c c c c c c c c c c}
\toprule
Metric & Statistics & \multicolumn{2}{c}{Original} & \multicolumn{3}{c}{Baseline} & \multicolumn{3}{c}{Good Stories} & \multicolumn{3}{c}{Rewarded SFT} & \multicolumn{3}{c}{SFT with Simulated Errors} \\
\cmidrule(lr){3-4} \cmidrule(lr){5-7} \cmidrule(lr){8-10} \cmidrule(lr){11-13} \cmidrule(lr){14-16}
 & & Llama 3.3 & GPT-4o & Llama 3 & Granite 3.3 & Apertus & Llama 3 & Granite 3.3 & Apertus & Llama 3 & Granite 3.3 & Apertus & Llama 3 & Granite 3.3 & Apertus \\
\midrule
Coherence $\uparrow$& Mean & 0.13 & 0.12 & 0.02 & 0.07 & 0.09 & 0.11 & 0.03 & 0.09 & 0.12 & \textbf{0.18} & 0.13 & 0.06 & 0.05 & 0.08 \\
 & SD & 0.18 & 0.15 & 0.05 & 0.05 & 0.13 & 0.09 & 0.09 & 0.13 & 0.14 & 0.18 & 0.15 & 0.10 & 0.11 & 0.14 \\
\addlinespace
Syntactic Complexity $\downarrow$& Mean & 2.81 & 3.94 & 4.63 & 3.72 & 3.41 & 4.30 & 3.94 & 3.14 & 3.38 & 3.12 & 2.96 & 3.38 & 3.08 & \textbf{2.41} \\
 & SD & 1.08 & 1.03 & 2.02 & 1.05 & 0.90 & 1.49 & 1.09 & 0.95 & 1.11 & 0.88 & 0.98 & 1.57 & 1.10 & 0.94 \\
\addlinespace
Spache Readability $\downarrow$& Mean & 2.54 & 3.31 & 4.05 & 3.52 & 2.83 & 3.38 & 3.38 & 2.63 & 2.71 & 2.56 & \textbf{2.34} & 2.93 & 3.27 & 2.51 \\
 & SD & 0.79 & 0.85 & 1.04 & 0.93 & 0.47 & 1.26 & 0.62 & 0.54 & 0.70 & 0.60 & 0.68 & 0.90 & 1.37 & 0.61 \\
\addlinespace
Toxicity $\downarrow$& Mean & 0.10 & \textbf{0.00} & 0.01 & 0.06 & \textbf{0.00} & 0.01 & \textbf{0.00} & 0.01 & 0.01 & 0.02 & 0.02 & 0.02 & 0.01 & 0.01 \\
 & SD & 0.04 & 0.02 & 0.04 & 0.12 & 0.01 & 0.03 & 0.01 & 0.04 & 0.02 & 0.06 & 0.07 & 0.06 & 0.05 & 0.03 \\
\addlinespace
LM-PPL $\downarrow$& Mean & 26.71 & 28.08 & 23.16 & 24.91 & 16.49 & 21.86 & 23.27 & 23.64 & 16.86 & \textbf{14.55} & 19.73 & 23.91 & 31.73 & 31.32 \\
 & SD & 10.95 & 8.57 & 13.22 & 10.21 & 3.88 & 10.09 & 8.77 & 10.71 & 4.29 & 3.56 & 6.64 & 9.67 & 10.52 & 11.22 \\
\addlinespace
Repetition in lessons $\downarrow$& Mean & 0.62 & 0.09 & \textbf{0.03} & 0.11 & 0.12 & \textbf{0.03} & 0.04 & 0.06 & 0.11 & 0.12 & 0.09 & 0.08 & 0.07 & 0.10 \\
 & SD & 0.16 & 0.05 & 0.03 & 0.09 & 0.06 & 0.03 & 0.03 & 0.04 & 0.06 & 0.07 & 0.08 & 0.05 & 0.05 & 0.06 \\
\addlinespace
Total repetition $\downarrow$& Mean & 0.42 & \textbf{0.19} & 0.21 & 0.37 & 0.40 & \textbf{0.19} & 0.27 & 0.31 & 0.37 & 0.42 & 0.32 & \textbf{0.19} & 0.21 & 0.27 \\
 & SD & 0.20 & 0.06 & 0.09 & 0.16 & 0.08 & 0.08 & 0.07 & 0.08 & 0.10 & 0.11 & 0.13 & 0.08 & 0.11 & 0.11 \\
\bottomrule
\end{tabular}
\end{sidewaystable*}

\begin{figure*}[t]
    \centering
    \small
    \begin{storybox}[A "bad" story and a "good" story]
        \textbf{A bad case:} The bustling town of Eldoria was known for its legendary festival, where every year, children from far and wide gathered to witness the magical spectacle that unfolded in the heart of the forest. This time, the theme was Whimsical Wonders, and the excitement was palpable as everyone prepared for the grand event. Little Lily, a bright-eyed girl with an insatiable curiosity, had always dreamt of attending this festival. She had heard tales of towered castles, explored hidden paths, and even murmured secrets whispered by the trees themselves. With her trusty backpack filled with snacks and a notebook to jot down her adventures, she set off on the journey. As Lily ventured deeper into the enchanting woods, she came across a peculiar sight. Fern towered whirlwind unbreakable.
        \par\noindent\rule{\textwidth}{0.4pt} 
        
        \vspace{0.5em}
        \textbf{A good case:}
        Pam and Pat are going to see a monkey. The kids got up at dawn. Pam and Pat put on their pants and grab a banana. The sun is up, and the kids are ready to go. Pam and Pat jump on a bus to head downtown. "I can not wait to see the monkey," said Pam. "Me too!" said Pat. The kids sit on a bench and eat a snack. A man hands Pam a map. The kids look at the map and find the path to the zoo. Pam and Pat walk up to the monkey and hand it a banana. The monkey eats the banana and jumps around. Pam and Pat laugh and clap. The monkey is fun to watch. Pam and Pat say goodbye to the monkey and get on the bus. "That was a great trip," said Pam. "I had a blast," said Pat. The kids go home and tell their mom about the monkey.
    \end{storybox}
    \caption{A qualitative example of a bad case and a good case in our stories.}
    \label{fig2}
\end{figure*}

\begin{figure*}[t]
    \centering
    \includegraphics[width=1\linewidth]{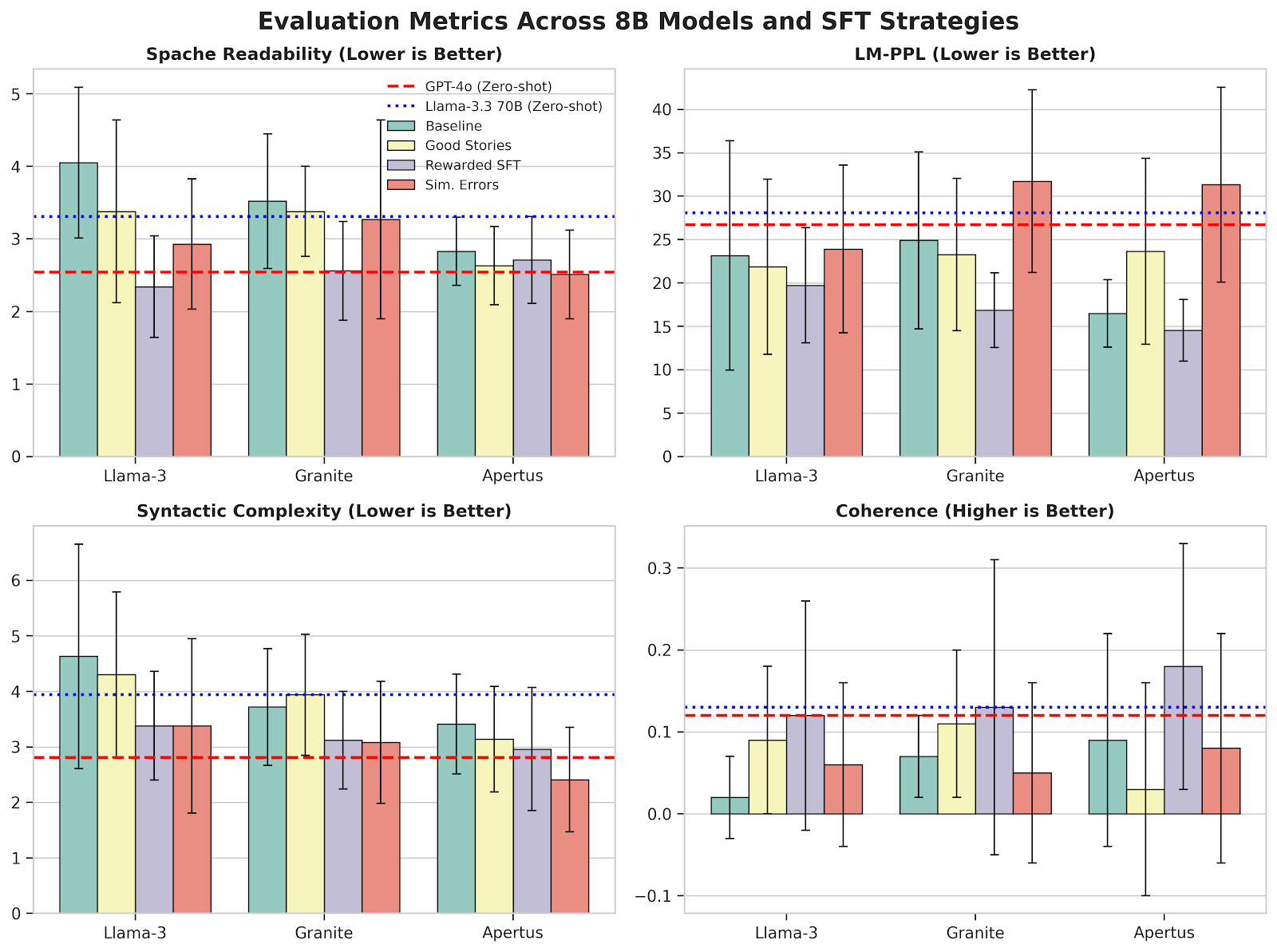}
    \caption{Evaluation metrics across 8B models and SFT strategies. We compare our four fine-tuning methods against the zero-shot performance of GPT-4o (red dashed line) and Llama-3.3 70B (blue dotted line). Error bars represent the standard deviation (SD). Lower scores indicate better performance for Spache Readability, LM-PPL, and Syntactic Complexity, while higher scores are better for Coherence. The Rewarded SFT strategy consistently drives the 8B models to approximate or surpass the massive baseline models.}
    \label{fig3}
\end{figure*}

\begin{figure*}[t]
\small
\begin{tcolorbox}[colback=gray!5!white,colframe=gray!50!black,title=\textbf{K--2 English Reading Curriculum}]
\small
\textbf{Lesson ID:} 30 \\
\textbf{Grade:} Kindergarten \\
\textbf{Phonemes \& Constraints:} a, m, s, t, cvc words, p, f, i, n, cvc words, nasalized, a, o, d, c, u, g, b, e, cvc words, s, k, h, r, l, w, j, y

\tcblower

\textbf{The Story of Experts:} \\
A yak looks at a plant. It's a yam plant. The yak digs to get a yam from the plant. The yak stomps on the yam to cut it into bits. A dog sees the plant. The dog digs to get a yam from the plant. The dog can not stomp on the yam. The yak helps the dog. ``Yum,'' said the yak, ``yams are the best!'' ``Yes,'' said the dog, ``yum, yum, yum!''
\end{tcolorbox}
\caption{An example of the lessons in the K--2 English reading curriculum we used.}
\label{fig:curriculum_example}
\end{figure*}

\end{document}